# A Two-Stage Cascade for Near-Real-Time Forest Anomaly Detection from Sentinel-1 SAR Time Series

Pann Thinzar Seint, Subas Chhatkuli, Bryan Atwood

*DAI Labs, K.K.*

Tropical forest monitoring is essential for global climate stability and biodiversity preservation. To address the urgent need for rapid, reliable detection of forest loss which is essential for timely intervention against illegal logging, supply chain transparency, land-use governance and carbon market standards, we introduce a two-stage statistics-encoder cascade for near-real-time anomaly detection using Sentinel-1 time series. Our system is designed to overcome two fundamental challenges in remote sensing: the cloud-cover limitations that restrict optical monitoring and seasonal backscatter variation that causes SAR systems to mistake natural moisture changes for forest loss. The architecture integrates two distinct analytical engines to ensure high-fidelity detection: (1) an adaptive, robust-statistics z-score test on co-registered Sentinel-1 VH backscatter, same-season historical baseline and (2) a learned confirmation gate based on the latent-space structural similarity (SSIM) of a convolutional autoencoder trained on stable-forest patches. A candidate disturbance is confirmed as an alert only when both stages agree, and is assigned a confidence score and a Low/Medium/High risk tier from its repeat-occurrence history. The system produces per-alert auditable confidence scores and area-in-hectares estimates directly compatible with Monitoring, Reporting and Verification (MRV) workflows, sustainable forestry management, operational field checks and environmental risk assessments. Beyond its primary application, the model's flexibility allows for critical environmental applications ranging from selective logging to large-scale agricultural encroachment mapping, flood mapping and so on.

***Keywords:*** *Sentinel-1; synthetic aperture radar; deforestation; SAR change detection; autoencoder; anomaly detection; REDD+; carbon MRV; near-real-time monitoring; tropical forest monitoring*

## 1. INTRODUCTION

Tropical forests cover approximately 2.5 billion hectares across Africa, Asia, and the Americas, storing an estimated 247 Pg C in aboveground biomass alone [1]. However, continued deforestation currently accounts for approximately 10% of annual anthropogenic greenhouse-gas (GHG) emissions, a magnitude comparable to the entire global transportation sector while simultaneously driving mass species extinction [2]. Given this, the ability to detect forest clearing with high spatial and temporal precision is critical. Ideally, alerts should be generated within days of a disturbance, as this timeline is a prerequisite for effective law enforcement and community-level intervention. Furthermore, such rapid detection is essential to meet the Monitoring, Reporting, and Verification (MRV) standards required by climate-finance frameworks, including REDD+ and the recently established Tropical Forests Forever Facility (TFFF) [3].

Current operational deforestation-alert systems such as the Global Land Analysis and Discovery (GLAD) alerts [4] largely rely on optical satellite imagery from platforms like Landsat-8/9 and Sentinel-2. While optical change detection is intuitive and well-validated, it is fundamentally limited by cloud cover. In humid tropical regions, it can take weeks to capture a single cloud-free image, allowing active clearing events to proceed undetected during these observation gaps. To address this, the European Space Agency's (ESA) Sentinel-1 has become the core of near-real-time monitoring systems. Using Sentinel-1 data, Reiche et al. [5] pioneered the the RAdar for Detecting Deforestation (RADD) system using a Bayesian probability framework while Doblas et al. [6] implemented the DETER-R system for the Brazilian Amazon using convolutional neural networks. Sentinel-1 provides freely available C-band Synthetic Aperture Radar (SAR) imagery with a 6-12 day revisit interval. Because SAR is an active sensor that transmits its own microwave pulses, it can penetrate cloud cover and operate regardless of sunlight, ensuring consistent monitoring in regions where optical sensors frequently fail.

While SAR technology overcomes cloud-cover limitations, it introduces unique technical challenges that optical sensors do not face. First, a coherent-imaging artifact creates pixel-level intensity variance that is independent of actual land-cover changes [7]. Second, seasonal and moisture-driven variations can alter VH backscatter signals, sometimes masking or mimicking actual structural canopy loss [8]. Critically, the sensor's acquisition geometry, including its incidence angle and ascending or descending pass direction, introduces systematic offsets in backscatter intensity. If monitoring images are compared against baselines drawn from different orbital geometries, these offsets can completely overwhelm the true anomaly signal. While recent deep-learning advances such as convolutional Bayesian fusion [9], and transformer-based architectures [10] have effectively addressed speckle and signal noise, the orbit-dependent viewing geometry variations (e.g., radar look-angle) remains unaddressed. Our findings indicate that geometry-driven bias is often the primary driver of false-positive alerts in large-scale operational deployments.

This paper introduces a two-stage cascade architecture designed through an interpretable framework. By integrating a transparent statistical detector with a learned structural confirmation gate, the approach balances model performance with operational rigor. The main contributions are:

- **(i)** A geometry-aware, adaptive-threshold anomaly detection stage featuring automated MAD-factor calibration.
- **(ii)** A convolutional autoencoder confirmation gate trained on stable-forest patches.
- **(iii)** Confidence scoring and risk-tier system optimized for operational Monitoring, Reporting, and Verification (MRV).
- **(iv)** A transferable architecture for environmental anomaly detection extensible beyond deforestation use cases.

## 2. RELATED WORK

The field of SAR-based forest monitoring has evolved significantly since the early JERS-1 and ERS missions, driven by the availability of dense, open-access Sentinel-1 time series has catalyzed a new generation of near-real-time (NRT) detection frameworks. For instance, Slagter et al. [11] demonstrated that fusing Sentinel-1 and Sentinel-2 data enables the classification of specific disturbance drivers, such as selective logging, road construction, and smallholder agriculture. Despite these advancements, a persistent operational challenge remains: balancing the desire for rapid detection ideally within one or two satellite revisit cycles against the need to suppress false positives, particularly in humid tropical environments where seasonal backscatter fluctuations often mimic structural forest loss.

Methods for SAR-based anomaly detection have transitioned from traditional statistical approaches, such as unsupervised Gaussianisation [12] and median absolute deviation (MAD) [13] to deep learning. Recently, autoencoders trained on stable forest patches have emerged as a powerful paradigm, as they flag anomalies based on deviations from learned representations [14]. This approach is particularly advantageous for tropical monitoring, where ground-truth labels are scarce and costly, as it eliminates the need for co-registered bitemporal image pairs.

Broadening the scope to multi-sensor approaches, Bullock et al. [15] achieved a peak producer's accuracy of 91.6% by fusing Landsat, Sentinel-2, and Sentinel-1 data, though their workflow involves a lag time of approximately 100 days. While self-supervised transformer architectures applied to fused time series have improved sensitivity to subtle degradation, they often require significantly higher computational resources than patch-level convolutional models. Finally, as emphasized in recent climate-finance literature [16], effective Monitoring, Reporting, and Verification (MRV) systems must provide more than just detection; they require auditable confidence metrics and area estimates aligned with REDD+ verification standards. Our proposed two-stage cascade is specifically designed to meet these operational requirements. Each alert is accompanied by an auditable confidence score and an area-in-hectares estimate aligned with reporting standards, making every detection independently traceable and verifiable rather than a simple binary flag. As a result, the method performs across a wide range of disturbance scales, from small 0.5-hectare events to large-scale deforestation.

# 3. METHODOLOGY

The system uses a two-stage process to monitor forests efficiently, as shown in **Figure 1**. First, a statistical layer quickly identifies potential changes. Then, a deep learning gate verifies these detections to remove false alerts. The following sections explain how each part of the system works.

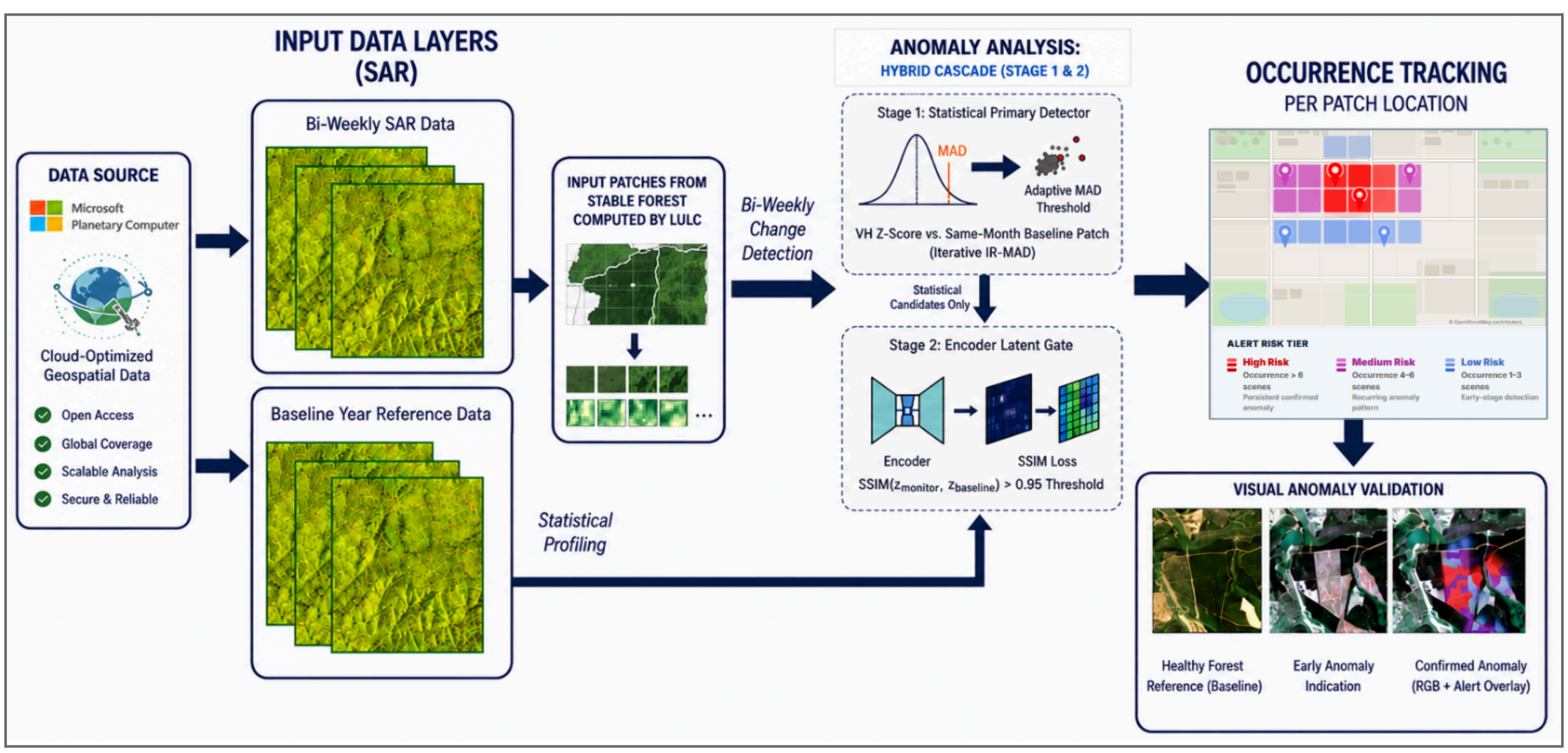

**Figure 1. System architecture for near-real-time forest disturbance detection**

### *3.1 Data Acquisition and Pre-Processing*

Sentinel-1 Radiometrically Terrain Corrected (RTC) scenes dual-polarisation VV+VH, Interferometric Wide (IW) swath mode are sourced from the Microsoft Planetary Computer STAC catalog. RTC products are pre-calibrated to ground-projected gamma-naught backscatter ($\gamma^0$), removing the need for a separate radiometric or terrain calibration step. All scenes are reprojected to a local UTM coordinate reference system (CRS) at 10 m resolution via bilinear resampling, and converted from linear-scale backscatter to decibels:

$$\gamma_dB = 10 \cdot \log_{10}(\gamma_linear) \quad (1)$$

Conversion to decibels is critical because the VH channel's speckle distribution is multiplicative in linear scale. All subsequent statistics are computed on γ_dB throughout the pipeline.

### *3.2 Stable-Forest Mask Construction*

To establish a reliable baseline, we define a static "stable-forest" domain. We generate binary stable-forest mask *M*, by intersecting Land Use/Land Cover (LULC) Maps from two distinct time $t_1$ and $t_2$ as shown in **Figure 2**. A pixel (*x*, *y*) is retained as stable forest only if it is classified as forest in both reference years:

$$M(x,y) = [\, LULC_\{t_1\}(x,y) = c_f \,] \cdot [\, LULC_\{t_2\}(x,y) = c_f \,] \quad (2)$$

*where LULC_t* denotes the LULC classification at reference year $t_1$ and year $t_2$, *c_f* is the forest class.

This mask restricts all subsequent change detection to land that was unambiguously forested prior to the monitoring period, suppressing alerts on pre-existing non-forest land uses and already-converted land. Monitoring and baseline scenes are both partitioned into an overlapping grid of $P \times P$ pixel patches ($P = 10$). A patch is retained as a candidate analysis unit only if at least a fraction $f_min \geq 0.90$ of its constituent pixels fall within ***M***.

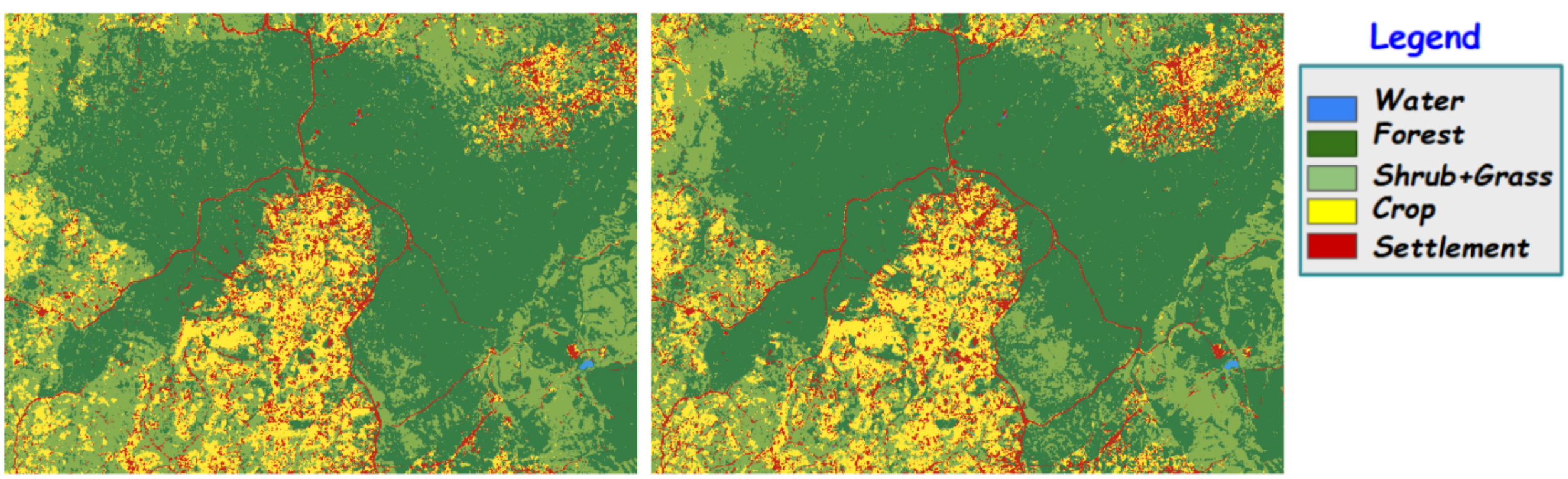

(a) LULC Maps at year t1 (b) LULC Maps at year t2

**Figure 2. Bi-temporal LULC classification Maps for stable-forest extraction**

### *3.3 Baseline Scene Selection and Orbit Geometry Matching*

To maintain operational integrity, the system enforces a strict geometric alignment protocol between the monitoring scene and its corresponding baseline. Because Sentinel-1 is a side-looking radar, comparing backscatter across divergent viewing geometries such as differing relative orbits or opposing (ascending versus descending) pass directions introduces systematic intensity offsets.

Two acquisitions sharing the same orbit number $r$ are acquired from the same incidence angle and look direction. The system selects, for each monitoring scene, the baseline candidate within a ±10 days calendar window that minimises. If no geometry-consistent baseline is available, the scene is skipped rather than compared against an inconsistent baseline.

### *3.4 Stage 1: Adaptive Statistical Anomaly Detection*

For each candidate patch, a per-patch z-score is computed by comparing the monitoring-period VH backscatter to the distribution of VH values in the corresponding baseline patch. Each patch is a 10*10 pixel window and patch coordinates are determined once from the stable-forest mask applied to the monitoring scene, producing a set of candidate locations ($i$, $j$). Critically, the identical set of coordinates is then used to extract the corresponding patch from the baseline scene at the same spatial position.

Let $\mu_mon$ denote the mean VH value (dB) of forest pixels in the monitoring patch, $\mu_base$ and $\sigma_base$ are the mean and standard deviation of the same-orbit baseline patch:

$$z = (\mu_mon - \mu_base) / \sigma_base \tag{3}$$

where $z$ is the standardized anomaly score, expressing how many baseline standard deviations the monitoring-period backscatter departs from the historical mean.

A strongly negative z-score indicates that monitoring-period backscatter is markedly lower than the historical baseline, the expected physical signature of canopy removal, which eliminates the volume scattering contribution of the forest canopy and typically reduces VH backscatter by 1.5 dB – 8 dB. Because the appropriate anomaly threshold varies by ROI and by season, a fixed global threshold would systematically over or under-alert across environments. Instead, the threshold $\tau$ is computed adaptively per scene using the Median Absolute Deviation (MAD) of all valid patch z-scores within the scene:

$$\tau = median(z) - k \cdot MAD(z) \tag{4}$$

where $\tau$ is the adaptive per-scene detection threshold and $MAD(z) = median(|z - median(z)|)$

The MAD is a robust scale estimator: unlike the standard deviation, it is resistant to the influence of extreme outliers (genuine disturbance patches) on the threshold calculation. The sensitivity factor $k$ is a per-ROI tunable parameter, increased during the wet season to compensate for elevated baseline variance from soil moisture. A patch is flagged at Stage 1 if:

$$z < \tau \tag{5}$$

#### *3.4.1 Automated MAD-Factor Calibration via Knee-Point Simulation*

Selecting an appropriate sensitivity factor $k$ is non-trivial: a value that is too low flags stable forest as anomalous, while a value that is too high suppresses genuine clearings. Rather than tuning $k$ by manual trial and error, the pipeline calibrates it offline through a knee-point simulation run once against a corpus of historical forest-patch z-scores ($N \geq 200$ patches).

For each candidate factor $k$ swept across the range [0.5, 3.0], a threshold is computed and applied to the historical z-score corpus, and the number of patches that would be flagged is recorded:

$$N(k) = \{patches : z < median(z) - k \cdot MAD(z)\} \quad (6)$$

This traces a monotonically decreasing sensitivity curve. The optimal factor $k^*$ is selected at the curve's geometric knee the point of maximum perpendicular distance from the chord connecting the swept curve's two endpoints:

Crucially, the iterative robust MAD procedure procedure first iteratively strips the extreme left tail of the z-score distribution, the genuine deforestation patches before estimating the stable-forest noise floor. This ensures the resulting sensitivity curve reflects the behaviour of the threshold on the stable-forest population alone, rather than being contaminated by the very disturbances the system is trying to detect.

Sweeping $k$ from low to high produces a monotonically decreasing detection curve: lenient thresholds flag many patches; strict thresholds flag few. The curve typically follows a steep initial descent that gradually flattens, indicating that beyond a certain point, increasing strictness yields rapidly diminishing reductions in flagged-patch count. The optimal factor $k^*$ is identified at the geometric knee of this curve, defined as the point of maximum perpendicular distance from the chord connecting the curve's two endpoints as shown in **Figure 3**:

$$k^* = argmax_k \ \ dist( (k, N(k)), \ \ chord[(k_min, N(k_min)) \rightarrow (k_max, N(k_max))] ) \quad (7)$$

where $N(k)$ is the number of patches flagged at threshold factor $k$; $(k_min, N(k_min)$ and $(k_max, N(k_max))$ are the two endpoints of the detection curve.

The knee represents the natural balance point between sensitivity and specificity for a given region because forest structure, terrain, and seasonal backscatter behaviour differ across regions, $k^*$ is calibrated independently per ROI rather than fixed globally.

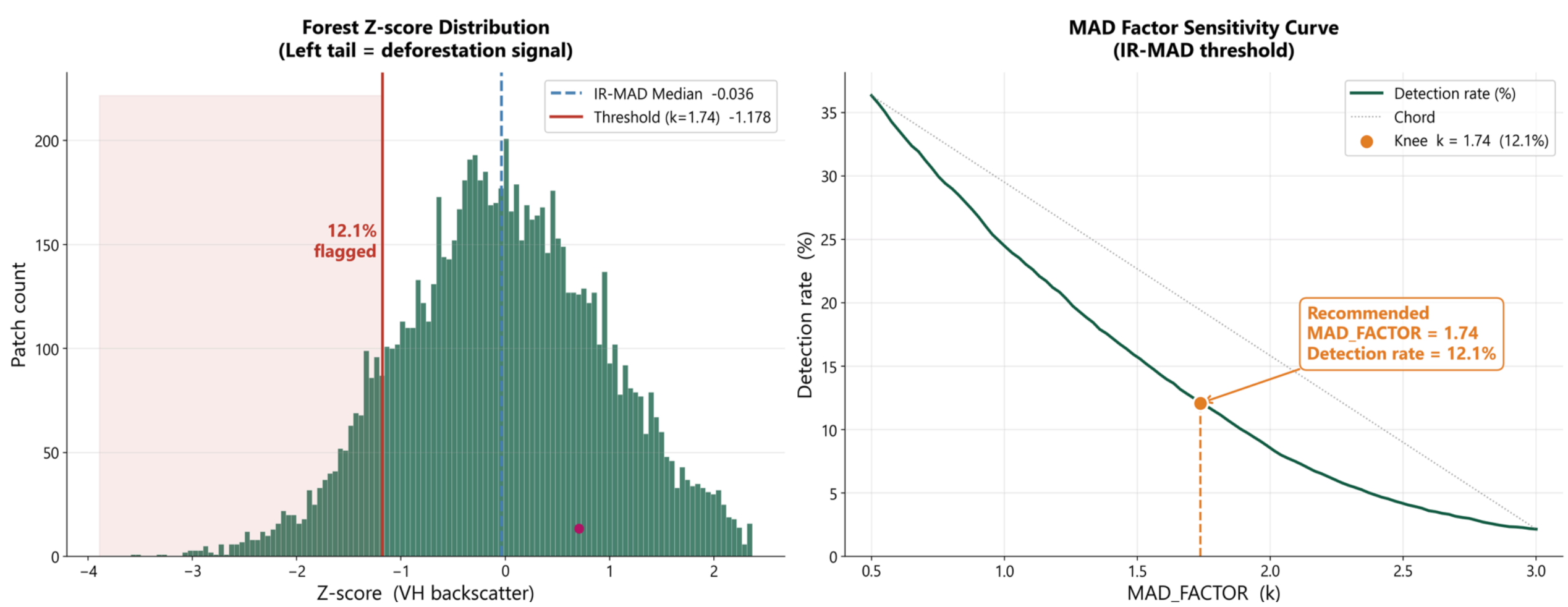

**Figure 3. MAD Sensitivity Simulation Curve**

### *3.5 Stage 2: Convolutional Autoencoder Confirmation Gate*

Patches flagged by Stage 1 are passed through a convolutional autoencoder *E*, previously trained on stable-forest patches using a reconstruction-loss objective. The encoder operates on 10×10-pixel, 3-channel (VV, VH, VV/VH-ratio) patches and consists of four convolutional blocks: a 3×3 convolution lifting the input to 32 channels at full resolution (10×10); a stride-2 3×3 convolution reducing spatial resolution to 5×5 while retaining 32 channels; a 3×3 convolution expanding to 64 channels at 5×5 (with 10% spatial dropout for regularisation); and a final stride-2 3×3 convolution producing a 32-channel, 3×3 bottleneck, a 288-dimensional latent representation. Each convolution is followed by Group Normalisation and a SiLU activation function. The decoder mirrors this structure and is used only during pre-training.

For every patch $\boldsymbol{x}$ flagged during the first stage, the system performs a structural consistency check by encoding both the monitoring patch, *x_mon*, and its co-located baseline, *x_base*, into a shared latent space. We then quantify the divergence between these latent representations using the Structural Similarity Index (SSIM) [17]:

$$s_recon_error = 1 - SSIM(\ E(x_mon),\ \ E(x_base)\ ) \qquad (8)$$

where *SSIM* is the structural similarity index, *E(x_mon), E(x_base)* are the latent representation of the monitoring patch and baseline patch after passing through the encoder.

In this framework, a higher reconstructed error score (*s_recon_error)* indicates that the encoder perceives a significant structural disparity between the two patches. To maintain high precision, we apply a thresholding mechanism where patches falling below the 10th percentile of reconstructed error score for the given scene are classified as "normal-looking" and subsequently discarded. Consequently, a candidate is only emitted as a confirmed alert if it successfully satisfies the requirements of both the statistical detector and the structural confirmation gate.

$$Alert(x) = [z(x) < \tau] \cdot [s_recon_error(x) > p_{10}(s_recon_error)] \qquad (9)$$

where $p_{10}$ *(s_recon_error)* denotes the 10th percentile of the reconstructed score distribution across the scene, i.e., the value below which the lowest-scoring 10% of patches fall.

The conjunction ensures that each stage's false-positive pathways, statistical outliers from speckle and seasonal noise on one hand, encoder false triggers from out-of-distribution patches on the other are independently filtered rather than compounded.

### *3.6 Confidence* Scoring

Alert patches are classified into a discrete risk tier by repeat-occurrence count *n*: Low ($1 \le n \le 3$) represented by blue color, Medium ($4 \le n \le 6$) represented by purple, and High ($n > 6$) represented by red as shown in **Figure 4**. A continuous risk surface is generated per monitoring date by seeding a Gaussian kernel. Outputs include a summary of alerted areas in hectares per risk tier.

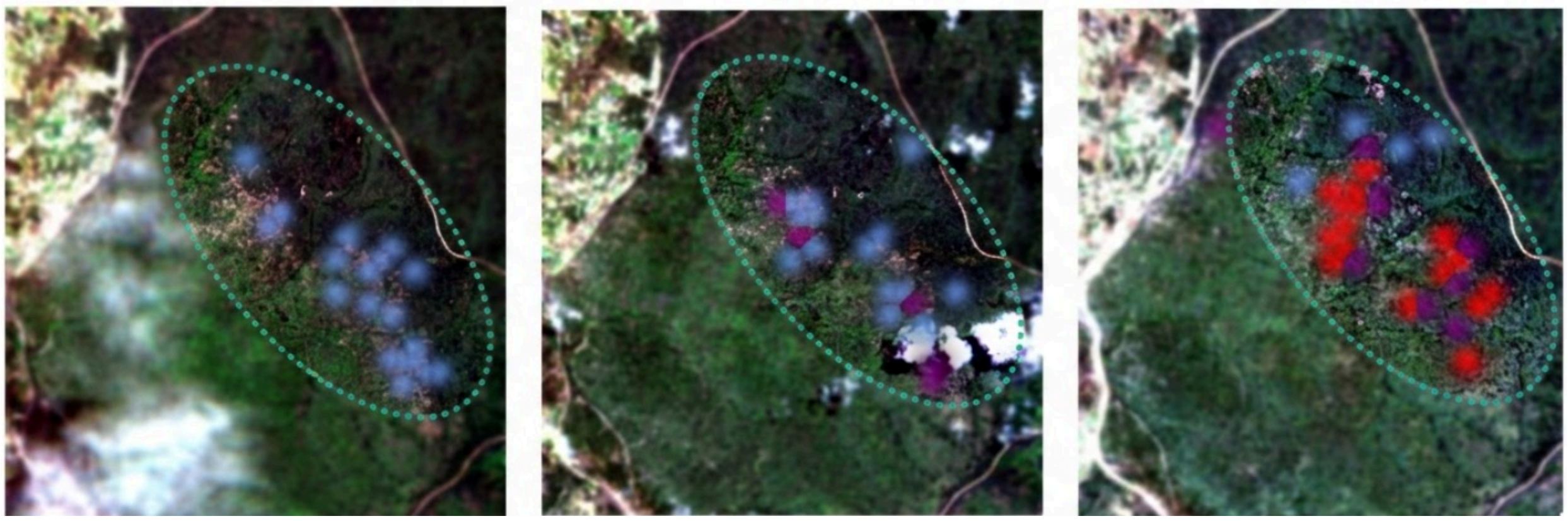

(a) (b) (c)

**Figure 4. System architecture for near-real-time forest disturbance detection: (a) Identification of early-stage deforestation (b) Monitoring of recurring deforestation events (c) Verification of high-risk forest loss and other medium and low-risk**

# 4. RESULTS AND DISCUSSION

## *4.1 Patch-Level Detection*

All detection operates at the patch level as a single spatial entity rather than individual pixels. This design choice directly addresses SAR's most fundamental noise source: speckle. By aggregating VH backscatter statistics across 10*10 forest pixels, the patch-level z-score averages out pixel-wise speckle variance, producing a far more stable anomaly signal than any pixel-wise approach could achieve. The patch also aligns naturally with the spatial scale of real deforestation events such as individual clearings, logging roads, and agricultural encroachments typically spanning areas far exceeding a single 10 m pixel, making it a physically meaningful detection unit. The minimum forest fraction threshold (90 %) further ensures that only patches with sufficient stable-forest coverage enter the pipeline, eliminating boundary ambiguity at the forest edge.

## *4.2 Effectiveness of the Two-Stage Conjunction*

Following geometry correction, the two-stage cascade consistently outperformed either component in isolation. Stage 1 alone still produced a residual false-positive load from wet-season moisture pulses that depressed VH backscatter coherently across entire patches, a spatially coherent signal that patch-level averaging cannot suppress. Stage 2 independently resolved these cases by verifying that the encoder's learned structural representation of the patch had genuinely changed, a check no statistical test can perform. The conjunction of both stages thus filtered two distinct and independent false-positive pathways simultaneously, without compounding their individual error rates, delivering the robustness of a learned model while preserving the full interpretability and auditability of the statistical detector.

### *4.3 Temporal Persistence*

A temporal-persistence mechanism was implemented and evaluated, requiring a candidate patch to be flagged across multiple consecutive scenes before a confirmed alert was issued. A patch undergoing progressive thinning from selective logging does not produce a single sharp z-score drop, it exhibits a z-score that drifts gradually negative over successive acquisitions, crossing the detection threshold inconsistently as moisture conditions and acquisition geometry fluctuate between scenes. A persistence window strict enough to eliminate isolated speckle spikes was also strict enough to silence this characteristic gradual drift, causing the system to miss precisely the low-intensity, incremental disturbances that are most difficult to detect by any other means and most in need of early intervention. The temporal persistence mechanism was therefore reverted in favour of per-scene independent patch-level detection. Temporal filtering is retained as an explicit, user-configurable post-processing step applied to the alert history database, where the sensitivity-specificity trade-off is fully visible.

## 5. Field Verification

We validated alert outputs from the anomaly detection pipeline using field data collected in Perekezi,Malawi in October 2025. GPS-tagged spot observations at candidate alert sites confirmed active deforestation, and overlaying these points with the corresponding alerts (**Figure 5**) shows spatial agreement between system-flagged patches and field conditions. This supports the validity of the per-scene detection approach described in Section 3.6 and section 4.3. Expanding ground-truth collection across wider geographic extents and disturbance types remains a critical objective for future work to validate detection metrics.

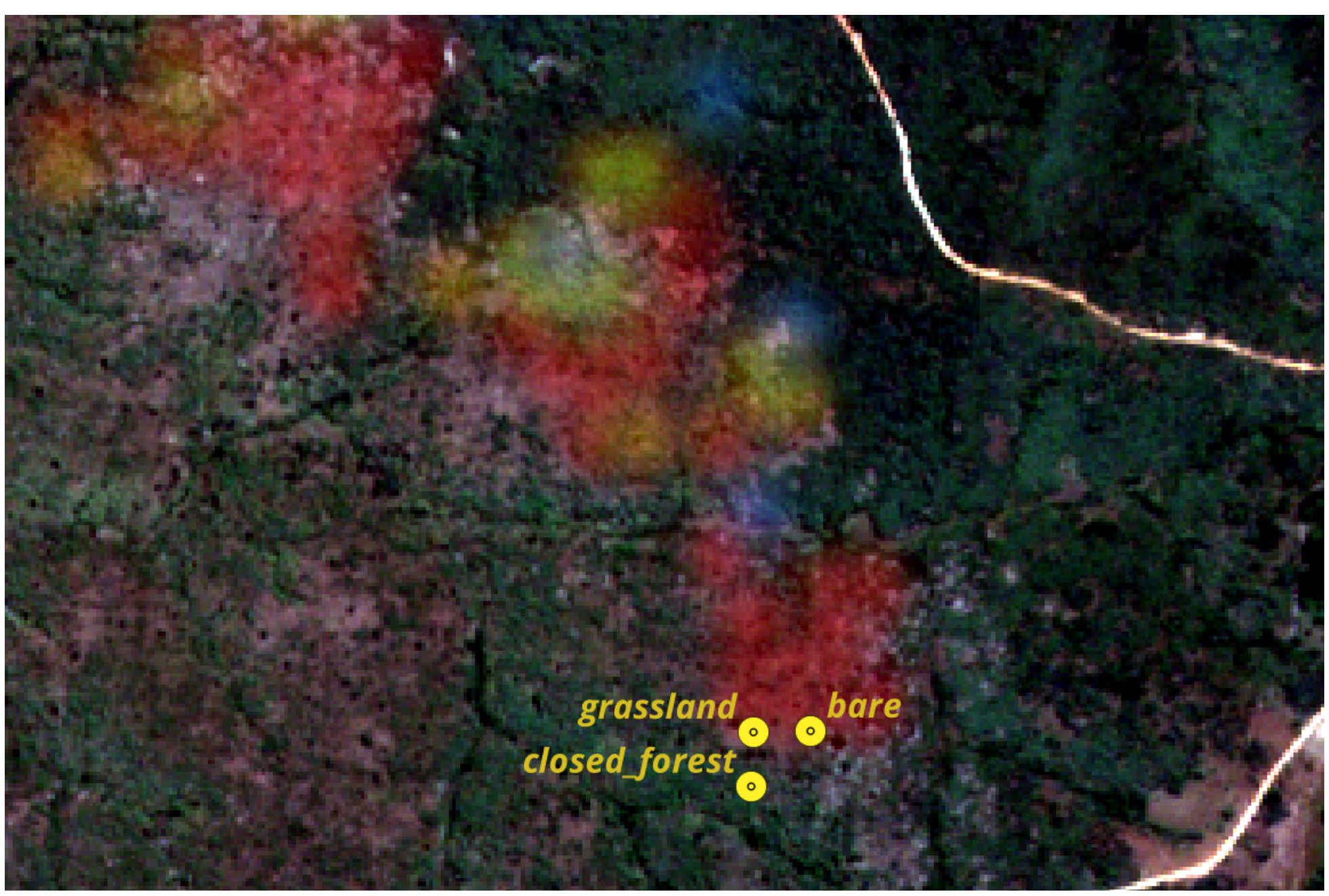

**Figure 5. Spatial overlay of system-generated deforestation alerts and GPS-tagged field verification points, Malawi, October 2025**

## 6. BROADER ENVIRONMENTAL IMPACT AND TRANSFERABILITY

Beyond tropical deforestation, the proposed architecture transfers directly to any monitoring problem where a structurally stable reference state can be defined. We outline the most direct transfer paths below, each representing a distinct avenue for future research and operational deployment.

- **Selective logging and forest degradation:** Sub-canopy structural loss that does not amount to full clear-cutting produces a smaller but physically analogous VH backscatter decrease. The cascade, retrained on a degradation-relevant baseline, extends naturally toward sub-canopy monitoring directly supporting REDD+ Tier 3 accounting, which requires emission factors for degradation separately from deforestation.
- **Illegal and artisanal small-scale mining:** Frequently occurring under forest canopy or in riparian zones poorly served by optical revisit, illegal mining produces tailings ponds and clearing signatures detectable in VH backscatter. The patch-level statistical/structural anomaly framework is directly applicable to mining-pit and corridor detection in the Peruvian Amazon, West African gold fields, and Southeast Asian forest concessions.
- **Flood extent and disaster response:** The same adaptive-thresholding and baseline-comparison machinery, applied to a flood-relevant backscatter signature rather than canopy loss, supports rapid-response inundation mapping unaffected by the storm-associated cloud that typically accompanies flood events directly applicable to Japan's post-Noto Peninsula earthquake response framework and to ASEAN disaster-management agencies.
- **Agricultural encroachment and illegal infrastructure:** Road-building and pipeline corridor incursion into protected-area forest produces a spatially elongated anomaly signature that the encoder gate's structural-similarity formulation is well positioned to discriminate from the more diffuse blob-like signature of clearing directly relevant to monitoring illegal solar farm expansion on protected agricultural land and to pipeline right-of-way surveillance in tropical concessions.
- **Carbon-market MRV at scale:** As voluntary and compliance carbon markets demand independently verifiable, near-real-time disturbance evidence for REDD+ and similar projects, the per-alert auditable confidence score, area-in-hectares estimate, and evidence layer produced by this system are directly positioned to support third-party verification workflows under the 2025 Tropical Forests Forever Facility and expanding J-Credit/Blue-Credit schemes across Asia-Pacific.

## 7. LIMITATIONS AND FUTURE WORK

The present system detects only backscatter-decrease disturbance signatures and does not address backscatter increases (e.g., secondary vegetation regrowth or wetland inundation). The MAD-factor calibration, while now automated via knee-point simulation, remains an offline, per-region step computed once against a historical z-score corpus rather than a continuously adaptive online parameter. The autoencoder is trained exclusively on stable-forest patches from the deployment ROI, which may limit zero-shot transferability to structurally dissimilar forest types such as dryland woodlands or mangroves without retraining. Validation in this study is operational and observational consistency of flagged-area trends, sanity-check warnings, and qualitative comparison against Sentinel-2 RGB composites rather than against an independently labelled ground-truth disturbance dataset; formal precision/recall evaluation against a curated benchmark is identified as primary future work.

Planned extensions include: (i) online MAD-factor adaptation using an exponential moving average of scene-level statistics; (ii) multi-modal fusion incorporating optical spectral bands (e.g., Red, NIR, and SWIR) alongside SAR polarizations (VH, VV, and VV/VH ratios) to improve classification between structural loss and non-forest land-cover transitions; (iii) cross-region encoder transfer experiments using contrastive pre-training on a global stable-forest patch corpus; and (iv) measuring detection reliability against verified disturbance records across the application domains.

# 8. CONCLUSIONS

This study has presented a two-stage statistics-encoder cascade for Sentinel-1 SAR-based, near-real-time forest anomaly detection, and the mathematical framework adaptive MAD-based z-score thresholding, automated knee-point sensitivity calibration, orbit-geometry-matched baseline selection, convolutional autoencoder latent-SSIM confirmation, and blended confidence scoring. The system overcomes two persistent barriers in operational forest monitoring: the cloud cover that blinds optical sensors for weeks at a time, and the seasonal backscatter variation that causes SAR systems to generate false alerts. The proposed system is interpretable: Stage 1 produces an auditable z-score and Stage 2 independently suppresses textural false triggers the statistical test cannot distinguish. Beyond deforestation, the architecture constitutes a transferable blueprint for cloud-independent environmental anomaly detection across forest degradation, illegal mining, wetland loss, flood response, agricultural encroachment, and carbon-market MRV applications of growing urgency under the 2025 Tropical Forests Forever Facility and expanding voluntary carbon markets across the Asia-Pacific region. As satellite revisit frequency increases and deep learning methods continue to mature, this kind of system is well-positioned to become the operational backbone of global forest monitoring.